\documentclass[manuscript,nonacm]{acmart}

\usepackage{multirow}
\usepackage{subcaption}

\title{ABCD: Alpha-Composited Block Coordinate Descent}

\author{Ka Heng Shiu}
\affiliation{%
  \institution{The University of Edinburgh}
  \city{Edinburgh}
  \country{United Kingdom}
}

\author{Kartic Subr}
\affiliation{%
  \institution{The University of Edinburgh}
  \city{Edinburgh}
  \country{United Kingdom}
}

\begin{document}

\begin{abstract}
We present ABCD (Alpha-Composited Block Coordinate Descent), an out-of-core training framework for alpha-composited radiance fields, instantiated here for 3D Gaussian Splatting. Our method reformulates training as block coordinate descent over spatial partitions: only one block of parameters is active at a time, while all others are frozen. By exploiting the associativity of alpha blending, these inactive regions can be pre-rendered and collapsed into foreground and background RGBA images.

As a result, for fixed partition size and image resolution, peak VRAM becomes $O(1)$ with respect to total scene extent, rather than growing with full scene size. This enables GPUs with limited memory to train scenes that would otherwise not fit in core. In experiments, our method closely preserves reconstruction quality of 3DGS, with less than 5\% PSNR degradation, while ABCD with compositing ablated suffers roughly 40\% degradation.
\end{abstract}

\maketitle

\begin{figure}[t]
    \centering
    \includegraphics[width=\linewidth]{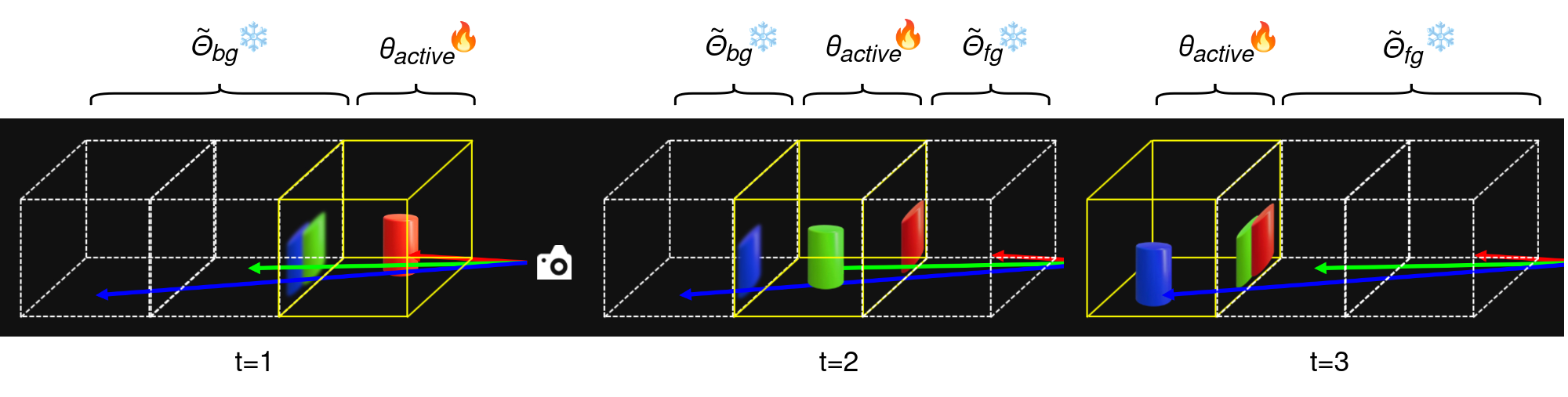}
    \caption{Overview of composited block descent with inactive regions cached as foreground and background RGBA images.}
    \Description{Diagram showing an active spatial partition rendered between cached foreground and background image layers.}
    \label{fig:overview}
\end{figure}

\section{Introduction}

Radiance fields are seeing increasing industry adoption, and their application is expanding to ever larger scenes. However, training such scenes quickly reaches the memory limits of modern GPUs, often requiring multi-GPU distributed setups that are prohibitively expensive for many users.

A natural first approach is to split a scene into smaller partitions, train them independently, and merge them afterward. However, NeRF-XL~\cite{li2024nerfxl} shows that this can introduce severe artifacts, since images frequently observe geometry spanning multiple partitions, causing each partition to compensate for missing content from the others. Works like BlockGaussians~\cite{wu2025blockgaussian} and Mega-NeRF~\cite{turki2022mega} improve the partition-train-merge pipeline through better partitioning, such as overlaps or auxiliary primitives near partition boundaries, but still depend on the quality of the decomposition. A LOD of Gaussians~\cite{windisch2025lod} reduces memory with a 3DGS-specific LOD tree, but these approaches are not general to arbitrary radiance field formulations.

To address this gap, we present ABCD, a principled out-of-core training method for alpha-composited radiance fields. Our formulation is block coordinate descent over spatial partitions~\cite{wright2015coordinate}, but exploits alpha compositing to pre-integrate all inactive parameters into foreground and background images. As a result, each update requires GPU memory only for the active partition and these two images. We demonstrate the method on 3D Gaussian Splatting~\cite{kerbl20233d}.

\section{Method}

Let the full scene parameters be partitioned into spatial blocks, $\Theta = \{\theta_i\}$ where each $\theta_i$ corresponds to a spatial partition $P_i$. At each stage of block coordinate descent, one block is selected as the active parameters $\theta_{\mathrm{active}}$, while all remaining parameters are held fixed.

As illustrated in Figure~\ref{fig:overview}, for a given active convex partition and camera view, the frozen parameters can be separated geometrically into those lying entirely in front of the active region and those lying entirely behind it. We denote these sets by $\Theta_{\mathrm{fg}}$ and $\Theta_{\mathrm{bg}}$, respectively. By associativity of alpha compositing and the fixed depth ordering induced by the active partition, their rendered contributions can be pre-accumulated into two flattened RGBA images: a foreground image $F$ and background image $B$. Optimizing $\theta_{\mathrm{active}}$ while compositing with $F$ and $B$ is equivalent to rendering with all frozen parameters explicitly present.

We first initialize scene primitives and divide them into convex spatial partitions (we use a regular 3D grid), yielding parameter blocks $\{\theta_i\}$. Next, for visible camera--partition pairs, we render and store compressed RGBA images on disk. To optimize a chosen active block $\theta_{\mathrm{active}}$, we gather cameras that observe its partition. For each such camera, cached renders from frozen partitions are composited into $F$ and $B$ according to depth ordering.

We then perform $T$ gradient steps on $\theta_{\mathrm{active}}$. At each step, a training camera observing the active partition is sampled, only the active partition is rendered, the result is composited with $F$ and $B$, reconstruction loss is evaluated, and gradients are backpropagated only through $\theta_{\mathrm{active}}$. After completing the $T$ updates, cached renders associated with that partition are refreshed, and optimization proceeds to the next block.

This produces a flexible memory hierarchy. Disk storage contains cached visible camera--partition renders. System memory may hold compressed cached images for faster reuse. GPU memory contains only $\theta_{\mathrm{active}}$ together with the required foreground/background images, rather than the full parameter set $\Theta$, and can be reduced further by streaming a single image pair at a time.

\section{Results}

We evaluate our method on two scenes from the original NeRF dataset~\cite{mildenhall2021nerf}, \textit{kitchen} and \textit{garden}, each containing approximately 200 training cameras. For both scenes, we partition the space using a regular grid with side length 5 units. We compare three training setups: standard 3D Gaussian Splatting (3DGS), ABCD with compositing ablated, and the full ABCD method. Although these scenes are relatively small, they represent a challenging setting for partition-based methods because spatial partitions frequently co-occur within the same camera views. ABCD with compositing ablated is therefore particularly susceptible to consistency artifacts, making this a stringent test of our compositing strategy.

\begin{figure}[t]
    \centering

    \begin{subfigure}[t]{0.32\linewidth}
        \centering
        \includegraphics[width=\linewidth]{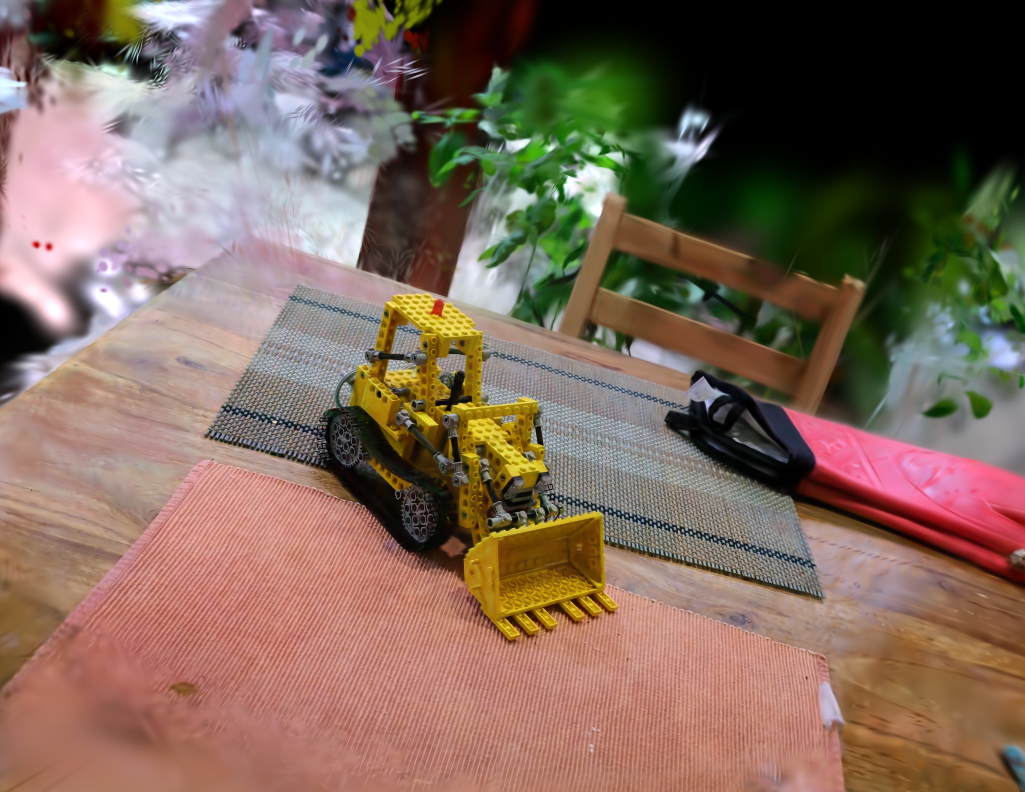}
        \caption{3DGS}
    \end{subfigure}
    \hfill
    \begin{subfigure}[t]{0.32\linewidth}
        \centering
        \includegraphics[width=\linewidth]{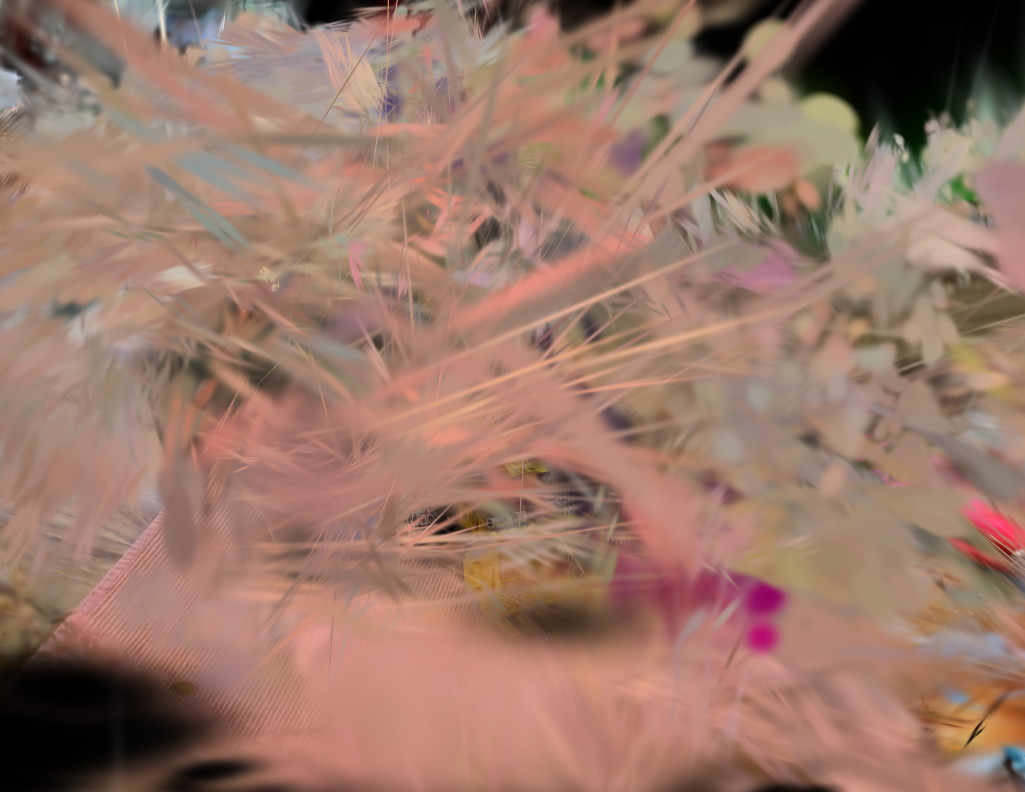}
        \caption{ABCD (ablated)}
    \end{subfigure}
    \hfill
    \begin{subfigure}[t]{0.32\linewidth}
        \centering
        \includegraphics[width=\linewidth]{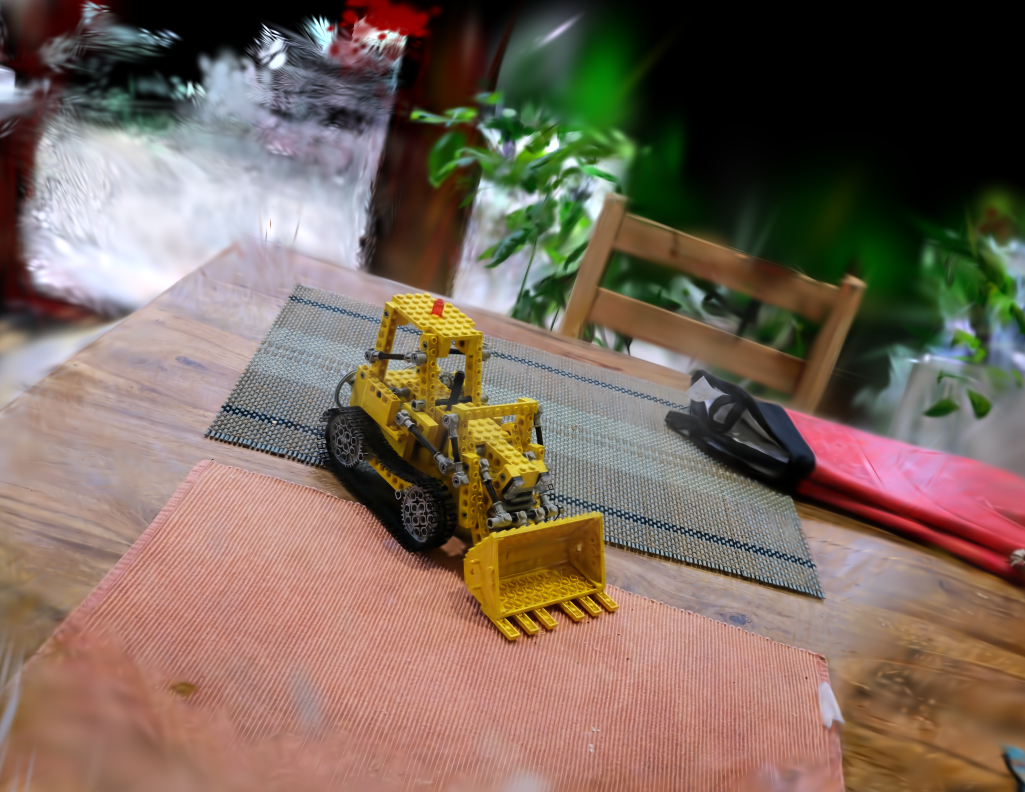}
        \caption{ABCD}
    \end{subfigure}

    \caption{Qualitative comparison on a held-out view. ABCD with compositing ablated introduces severe boundary and consistency artifacts, while the full ABCD method closely matches the visual quality of 3DGS.}
    \Description{Three rendered views comparing 3D Gaussian Splatting, ABCD with compositing ablated, and ABCD.}
    \label{fig:qualitative_comparison}
\end{figure}

\begin{figure}[t]
    \centering

    \begin{subfigure}[t]{0.49\linewidth}
        \centering
        \includegraphics[width=\linewidth]{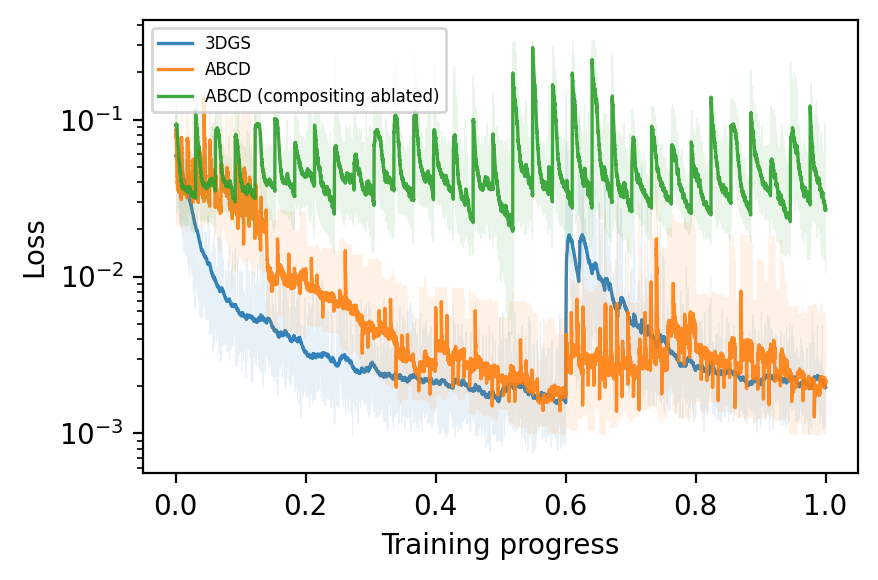}
        \caption{Training loss trajectories.}
    \end{subfigure}
    \hfill
    \begin{subfigure}[t]{0.49\linewidth}
        \centering
        \includegraphics[width=\linewidth]{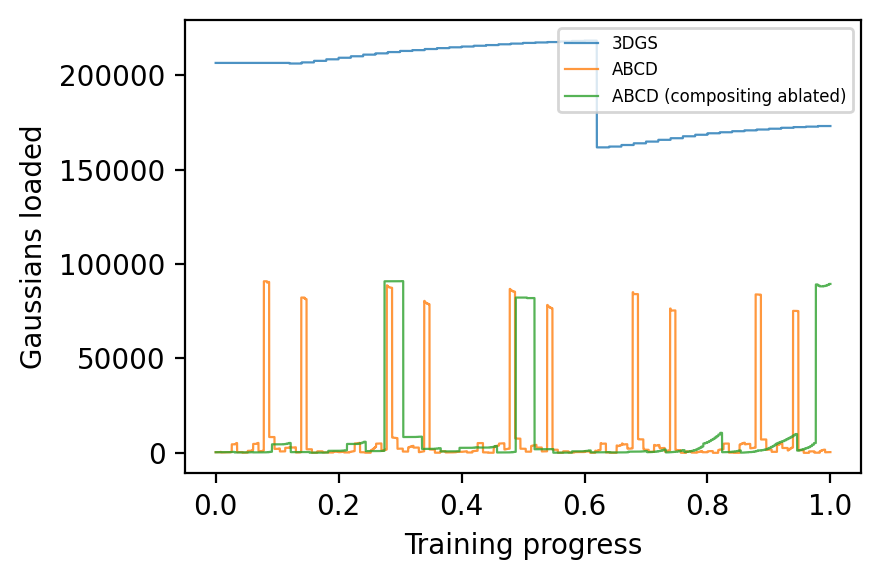}
        \caption{Active Gaussians resident in GPU memory.}
    \end{subfigure}

    \caption{Training dynamics of 3DGS, ABCD with compositing ablated, and ABCD. Left: ABCD follows a loss trajectory similar to 3DGS, whereas removing compositing substantially impairs optimization. Right: ABCD retains the low active Gaussian residency of partitioned training, demonstrating that it preserves the memory behaviour of ABCD with compositing ablated while recovering most of the lost reconstruction quality.}
    \Description{Training losses and active Gaussian counts for 3DGS, ABCD with compositing ablated, and ABCD.}
    \label{fig:training_dynamics}
\end{figure}

\begin{table}[t]
  \centering
  \caption{Comparison across methods on \textit{garden} and \textit{kitchen} scenes.}
  \label{tab:results}
  \begin{tabular}{p{2.2cm} p{2.8cm} *{5}{p{1.6cm}}}
    \toprule
    \textbf{Dataset} & \textbf{Method} & \textbf{VRAM (GB) $\downarrow$} & \textbf{RAM (GB) $\downarrow$} & \textbf{Time (s) $\downarrow$} & \textbf{PSNR $\uparrow$} & \textbf{SSIM $\uparrow$} \\
    \midrule

    \multirow{3}{*}{Garden}
    & 3DGS & 0.88 & 7.88  & 99.6   & 24.01 & 0.596 \\
    & ABCD (ablated) & 0.69 & 8.98  & 402.7  & 16.56 & 0.356 \\
    & ABCD & 0.80 & 25.02 & 1081.3 & 23.65 & 0.575 \\

    \midrule

    \multirow{3}{*}{Kitchen}
    & 3DGS & 0.97 & 3.73  & 67.4   & 27.03 & 0.861 \\
    & ABCD (ablated) & 0.75 & 5.60  & 441.3  & 15.64 & 0.504 \\
    & ABCD & 0.77 & 17.73 & 1041.0 & 25.80 & 0.839 \\

    \bottomrule
  \end{tabular}
\end{table}

We treat 3DGS as the quality reference, and compare the full ABCD method against ABCD with compositing ablated. Averaged across both scenes, ABCD with compositing ablated reduces PSNR from 25.52 to 16.10, a 36.9\% degradation relative to the reference. ABCD improves this substantially, reaching an average PSNR of 24.73, which is only a 3.1\% drop from 3DGS. The same trend appears in SSIM: ABCD with compositing ablated reduces average SSIM from 0.728 to 0.430, a 41.0\% drop, whereas ABCD reaches 0.707, only 3.0\% below the reference.

Figure~\ref{fig:training_dynamics} further illustrates the optimization behaviour of the three methods. ABCD follows a training trajectory similar to 3DGS, whereas ablating compositing substantially degrades optimization. Despite recovering most of the reconstruction quality lost by ABCD with compositing ablated, ABCD retains nearly identical active Gaussian residency within GPU memory, preserving the principal memory advantages of partitioned training.

The more significant result is memory scaling. In conventional training, VRAM grows with total scene size because all parameters must remain resident during optimization. Under ABCD, VRAM depends only on $\theta_{\mathrm{active}}$ together with one foreground/background image pair for the sampled view, making memory effectively independent of total scene extent for fixed partition size and image resolution. Averaged across scenes, peak VRAM is 0.785\,GB for ABCD, compared with 0.720\,GB for ABCD with compositing ablated and 0.925\,GB for 3DGS. Thus, relative to ABCD with compositing ablated, our method increases VRAM by only 0.065\,GB on average, or 9.0\%, while still remaining 15.1\% below 3DGS. While these reductions appear modest, the active Gaussian counts in Figure~\ref{fig:training_dynamics} show that ABCD preserves the same partitioned memory behaviour as ABCD with compositing ablated. The asymptotic benefits are masked in these experiments because the scenes contain relatively few Gaussians, causing optimizer state and image-related overheads to dominate total GPU memory consumption.

System RAM usage is higher: averaged across scenes, ABCD uses 21.38\,GB, compared with 7.29\,GB for ABCD with compositing ablated and 5.81\,GB for 3DGS. Relative to ABCD with compositing ablated, this is an increase of 14.09\,GB, or 193.2\%. This additional memory stores cached rendered images required by cameras that observe the active partition after frustum filtering. These images are accessed only when switching active partitions; many optimization steps are then performed on the same shard, so transfer cost is strongly amortized. If desired, the cache can therefore be stored on disk and streamed with limited impact on total training time.

This memory tradeoff also incurs additional runtime. Averaged across scenes, training time increases from 83.5\,s for 3DGS and 422.0\,s for ABCD with compositing ablated to 1061.2\,s for ABCD. Relative to ABCD with compositing ablated, this is an increase of 639.2\,s, or 151.5\%. We attribute this slowdown primarily to reduced rasterizer occupancy: active partitions contain relatively few Gaussians, making kernel launch overheads more significant and preventing efficient utilization of the rendering pipeline. We expect this effect to diminish as scene scale increases and active partitions contain more primitives. More importantly, ABCD targets settings in which conventional training may be infeasible on the available hardware, making increased runtime a practical trade-off for enabling reconstruction altogether.

Overall, our method shifts the limiting resource for large-scene training from scarce GPU memory to cheaper storage tiers, while preserving nearly the reconstruction quality of unmodified 3DGS.

\section{Limitations and Future Work}

Our current experiments are limited by time and engineering effort. In particular, we did not fully tune densification and training hyperparameters for 3D Gaussian Splatting, and evaluated only relatively small scenes containing modest numbers of Gaussians. Consequently, optimizer state, image storage, and rasterization overheads constitute a substantial fraction of total GPU memory usage, masking the asymptotic memory savings of ABCD. We therefore expect the advantages of our method to become substantially more pronounced on larger real-world scenes where Gaussian parameters dominate memory consumption.

A practical limitation is that training a partition requires loading cached render pairs for cameras that observe that partition. With only frustum culling, this quantity can still grow with scene scale, although more slowly than storing the full model. In practice, distant partitions eventually occupy negligible image area and can be ignored, while many others become occluded. Incorporating visibility thresholds and dynamic occlusion culling should further reduce memory and storage costs.

Theoretical convergence behavior also remains open. Our method follows block coordinate descent, which is often effective for smooth objectives~\cite{wright2015coordinate}, but it would be valuable to better understand its behavior for Gaussian Splatting specifically, especially when variables are grouped spatially into grid cells.

Our current Gaussian assignment is also approximate: Gaussians are assigned to partitions using their means, even though their covariance may cross partition boundaries. Since strict partition membership determines correctness of the foreground/background decomposition, more careful assignment rules or overlapping boundary regions could further improve robustness.

More broadly, partitioned processing of radiance fields appears promising beyond out-of-core training. While we focus on single-GPU block coordinate descent, future work should evaluate ABCD on substantially larger real-world datasets to validate its asymptotic memory advantages and practical scalability. Distributed variants could exchange only rendered partition images intermittently across compute nodes, resembling asynchronous stochastic optimization with delayed updates and potentially enabling decentralized or bandwidth-limited large-scale Gaussian Splatting training.

\bibliographystyle{ACM-Reference-Format}
\bibliography{refs.bib}

\end{document}